# Artificial Intelligence and Legal Liability

J.K.C. Kingston[1]

**Abstract**  A recent issue of a popular computing journal asked which laws would apply if a self-driving car killed a pedestrian. This paper considers the question of legal liability for artificially intelligent computer systems. It discusses whether criminal liability could ever apply; to whom it might apply; and, under civil law, whether an AI program is a product that is subject to product design legislation or a service to which the tort of negligence applies. The issue of sales warranties is also considered.

A discussion of some of the practical limitations that AI systems are subject to is also included.

## 1 Introduction

A recent issue of a popular computing journal [1] posed the following question:

"It is the year 2023, and for the first time, a self-driving car navigating city streets strikes and kills a pedestrian. A lawsuit is sure to follow. But exactly which laws will apply? No-one knows."

The article goes on to suggest that the laws that are likely to apply are those that deal with products with a faulty design. However, it argues that following this legal route holds back the development of self-driving cars, as settlements for product design cases (in the USA) are typically almost ten times higher than for cases involving human negligence, and that does not include the extra costs associated with product recalls to fix the issue. It goes on to argue that such cases should instead be dealt with as cases of negligence, just as they would for a human driver; the author points out that a standard handbook of US tort law [2] states that "A bad state of mind is neither necessary nor sufficient to show negligence; conduct is everything."

It may be that the issue will arise even sooner than the year 2023. The author of this paper recently hired a car that included several new safety features. One of these features was that, if the car's radars detected an imminent collision while the car was travelling at between 4 and 19 mph, the car's engine would cut out to help prevent the collision.

1 University of Brighton, BN2 4JG, UK
j.k.kingston@brighton.ac.uk

While reversing the car out of a driveway, the author drove too close to a hedge. The car sounded its proximity alarm, and cut the engine. However, even when the steering wheel was turned so that the car would miss the hedge, the engine would not restart while the car was in reverse gear. The author had to put the car into a forward gear and travel forward slightly before he was able to continue reversing out.

All of this took place wholly within the driveway. However, if it had taken place while the rear end of the car was projecting into the road, with a heavy lorry travelling at some speed towards the car, most drivers would prefer to risk getting their paintwork scratched by a hedge than sitting in a car that refused to restart and complete the desired manoeuvre. It seems inevitable that soon, some driver will blame these 'safety features' for their involvement in a serious accident.

The purpose of this paper is to consider the current capabilities, or lack of them, of artificial intelligence, and then to re-visit the question of where legal liability might lie in the above cases.

First, it is important to establish what this paper means by the term "artificial intelligence". There are researchers in the AI field who consider anything that mimics human intelligence, by whatever method, to be "artificial intelligence"; there are others who think that the only "artificially intelligent" programs are those that mimic the way in which humans think. There are also those in the field of information systems who would classify many "artificially intelligent" programs as being complex information systems, with 'true' artificial intelligence being reserved for the meta-level decision making that is sometimes characterised as 'wisdom'.

In this paper, any computer system that is able to recognise a situation or event, and to take a decision of the form "IF this situation exists THEN recommend or take this action" is taken to be an artificially intelligent system.

## 2 Legal Liability

### 2.1 Criminal Liability

The references cited below refer primarily to US law; however, many other jurisdictions have similar legislation in the relevant areas.

In [3], Gabriel Hallevy discusses how, and whether, artificial intelligent entities might be held criminally liable. Criminal laws normally require both an *actus reus* (an action) and a *mens rea* (a mental intent), and Hallevy helpfully classifies laws as follows:

1. Those where the *actus reus* consists of an action, and those where the *actus reus* consists of a failure to act;
2. Those where the *mens rea* requires knowledge or being informed; those where the *mens rea* requires only negligence ("a reasonable person would have known"); and strict liability offences, for which no *mens rea* needs to be demonstrated.

Hallevy goes on to propose three legal models by which offences committed by AI systems might be considered:

1. Perpetrator-via-another. If an offence is committed by a mentally deficient person, a child or an animal, then the perpetrator is held to be an innocent agent because they lack the mental capacity to form a *mens rea* (this is true even for strict liability offences). However, if the innocent agent was instructed by another person (for example, if the owner of a dog instructed his dog to attack somebody), then the instructor is held criminally liable (see [4] for US case law).

   According to this model, AI programs could be held to be an innocent agent, with either the software programmer or the user being held to be the perpetrator-via-another.

2. Natural-probable-consequence. In this model, part of the AI program which was intended for good purposes is activated inappropriately and performs a criminal action. Hallevy gives an example (quoted from [5]) in which a Japanese employee of a motorcycle factory was killed by an artificially intelligent robot working near him. The robot erroneously identified the employee as a threat to its mission, and calculated that the most efficient way to eliminate this threat was by pushing him into an adjacent operating machine. Using its very powerful hydraulic arm, the robot smashed the surprised worker into the machine, killing him instantly, and then resumed its duties.

   The normal legal use of "natural or probable consequence" liability is to prosecute accomplices to a crime. If no conspiracy can be demonstrated, it is still possible (in US law) to find an accomplice legally liable if the criminal acts of the perpetrator were a natural or probable consequence (the phrase originated in [6]) of a scheme that the accomplice encouraged or aided [7], as long as the accomplice was aware that some criminal scheme was under way.

So users or (more probably) programmers might be held legally liable if they knew that a criminal offence was a natural, probable consequence of their programs/use of an application. The application of this principle must, however, distinguish between AI programs that 'know' that a criminal scheme is under way (i.e. they have been programmed to perform a criminal scheme) and those that do not (they were programmed for another purpose). It may well be that crimes where the *mens rea* requires knowledge cannot be prosecuted for the latter group of programs (but those with a 'reasonable person' *mens rea*, or strict liability offences, can).

3. Direct liability. This model attributes both *actus reus* and *mens rea* to an AI system.

   It is relatively simple to attribute an *actus reus* to an AI system. If a system takes an action that results in a criminal act, or fails to take an action when there is a duty to act, then the *actus reus* of an offence has occurred.

   Assigning a *mens rea* is much harder, and so it is here that the three levels of *mens rea* become important. For strict liability offences, where no intent to commit an offence is required, it may indeed be possible to hold AI programs criminally liable. Considering the example of self-driving cars, speeding is a strict liability offence; so according to Hallevy, if a self-driving car was found to be breaking the speed limit for the road it is on, the law may well assign criminal liability to the AI program that was driving the car at that time.

   This possibility raises a number of other issues that Hallevy touches on, including defences (could a program that is malfunctioning claim a defence similar to the human defence of insanity? Or if it is affected by an electronic virus, could it claim defences similar to coercion or intoxication?); and punishment (who or what would be punished for an offence for which an AI system was directly liable?).

## 2.2 The Trojan defence

In the context of defences against liability for AI systems, it is important to mention a number of cases where a defendant accused of cybercrime offences has successfully offered the defence that his computer had been taken over by a Trojan or similar malware program, which was committing offences using the defendant's computer but without the defendant's knowledge. A review of such cases can be found in [8]; they include a case in the United Kingdom when a computer containing indecent pictures of children was also found to have eleven Trojan programs on it, and another UK case where a teenage computer hacker's defence to a charge of executing a denial of service attack was that the attack had been performed from the defendant's computer by a Trojan program, which had subsequently wiped itself from the computer before it was forensically analysed. The defendant's lawyer successfully convinced the jury that such a scenario was not beyond reasonable doubt.

## 2.3 Civil Law: Torts and Breach of Warranty

### 2.3.1 Negligence

When software is defective, or when a party is injured as a result of using software, the resulting legal proceedings normally allege the tort of negligence rather than criminal liability [9]. Gerstner [10] discusses the three elements that must normally be demonstrated for a negligence claim to prevail:
1. The defendant had a duty of care;
2. The defendant breached that duty;
3. That breach caused an injury to the plaintiff.

Regarding point 1, Gerstner suggests there is little question that a software vendor owes a duty of care to the customer, but it is difficult to decide what standard of care is owed. If the system involved is an "expert system", then Gerstner suggests that the appropriate standard of care is that of an expert, or at least of a professional.

On point 2, Gerstner suggests numerous ways in which an AI system could breach the duty of care: errors in the program's function that could have been detected by the developer; an incorrect or inadequate knowledge base; incorrect or inadequate documentation or warnings; not keeping the knowledge up to date; the user supplying faulty input; the user relying unduly on the output; or using the program for an incorrect purpose.

As for point 3, the question of whether an AI system can be deemed to have caused an injury is also open to debate. The key question is perhaps whether the AI system *recommends* an action in a given situation (as many expert systems do), or *takes* an action (as self-driving and safety-equipped cars do). In the former case, there must be at least one other agent involved, and so causation is hard to prove; in the latter case, it is much easier.

Gerstner also discussed an exception under US law for "strict liability negligence." This applies to products that are defective or unreasonably dangerous when used in a normal, intended or reasonably foreseeable manner, and which cause injury (as opposed to economic loss). She discusses whether software is indeed a 'product' or merely a 'service'; she quotes a case in which electricity was held to be a product [11], and therefore leans towards defining software as a product rather than a service. Assuming that software is indeed a product, it becomes incumbent on the developers of AI systems to ensure that their systems are free from design defects; manufacturing defects; or inadequate warning or instructions.

Cole [12] provides a longer discussion of the question of whether software is a product or a service. His conclusion is that treating AI systems as products is "partially applicable at best", and prefers to view AI as a service rather than a product; but he acknowledges that law in the area is ill-defined.

Cole cites some case law regarding the "duty of care" that AI systems must abide by:

1. In [14], a school district brought a negligence claim against a statistical bureau that (allegedly) provided inaccurate calculations of the value of a school that had burned down, causing the school district to suffer an underinsured loss. The duty being considered was the duty to provide information with reasonable care. The court considered factors including: the existence, if any, of a guarantee of correctness; the defendant's knowledge that the plaintiff would rely on the information; the restriction of potential liability to a small group; the absence of proof of any correction once discovered; the undesirability of requiring an innocent party to carry the burden of another's professional mistakes; and, the promotion of cautionary techniques among the informational (tool) providers.

2. Based on [15], Cole discusses the duty to provide reasonable conclusions from unreasonable inputs. He follows [16] to suggest that AI developers probably have an affirmative duty to provide relatively inexpensive, harmless, and simple, input error-checking techniques, but notes that these rules may not apply where the AI program is performing a function in which mistakes in input may be directly life-threatening (e.g. administering medicine to a patient); in such cases, he suggests applying the rules relating to "ultra-hazardous activities and instrumentalities" instead [17].

3. Cole suggests that AI systems must be aware of their limitations, and this information must be communicated to the purchaser. It is well established that vendors have a duty to tell purchasers of any known flaws; but how can unknown weaknesses or flaws be established, and then communicated?

**2.3.2 Breach of Warranty**

If an AI system is indeed a product, then it must be sold with a warranty; even if there is no express warranty given by the vendor (or purchased by the user), there is an implied warranty that it is (to use the phrase from the UK Sale of Goods Act 1979), "satisfactory as described & fit for a reasonable time." Some jurisdictions permit implied warranties to be voided by clauses in the contract; however, when an AI system is purchased built into other goods (such as a car), it seems unlikely that any such contractual exclusions (e.g. between the manufacturer of the car and the supplier of the AI software) could successfully be passed on to the purchaser of the car.

## *2.3 Legal liability: Summary*

So it seems that the question of whether AI systems can be held legally liable depends on at least three factors:
- The limitations of AI systems, and whether these are known and communicated to the purchaser;
- Whether an AI system is a product or a service;
- Whether the offence requires a *mens rea* or is a strict liability offence.

If an AI system is held liable, the question arises of whether it should be held liable as an innocent agent, an accomplice, or a perpetrator.

The final section of this paper consider the first of these three factors.

## 3 Limitations of AI systems

The various limitations that AI systems are subject to can be divided into two categories:
- Limitations that human experts with the same knowledge are also subject to;
- Limitations of artificial intelligence technology compared with humans.

### *3.1 Limitations that affect both AI systems and human experts*

The limitations that affect both AI systems and human experts are connected with the knowledge that is specific to the problem.

Firstly, the knowledge may change very rapidly. This requires humans and AI systems both to know what the latest knowledge is, and also to identify which parts of their previous knowledge is out of date. Whether this is an issue depends almost entirely on the domain: in our example of automated car driving, the knowledge that is required to drive a car changes very slowly indeed. However, in the world of cyber security, knowledge of exploits and patches changes on a daily basis.

Secondly, the knowledge may be too vast for all possibilities to be considered. AI systems can actually perform better than human experts at such tasks – it is feasible to search thousands, or even hundreds of thousands of solutions – but there are still some tasks where the scope is even wider than that. This is typically the case in planning and design tasks, where the number of possible plans or designs may be close to infinite. (In contrast, scheduling and configuration, which require planning and design within a fixed framework, are less complex, though the possible options may still run into thousands). In such cases, AI systems can promise to give a good answer in most cases, but cannot guarantee that they will give the best answer in all cases.

From a legal standpoint, it could be argued that the solution to such issues is for the vendor to warn the purchaser of an AI system of these limitations. In fast-changing domains, it may also be considered legally unreasonable if the vendor does not provide a method for frequently updating the system's knowledge. This raises the question of where the boundaries of 'fast-changing' lie. As ever, the legal test is reasonableness, which is usually compared against the expected life of an AI system; so if knowledge was expected to change annually (e.g. in an AI system for calculating personal tax liability), then it would probably be judged reasonable for a vendor to warn that the knowledge was subject to change. However, it would probably not be judged 'reasonable' for the vendor to provide automatic updates to the knowledge, because the complexity of tax law is such that any updates would not merely require downloading files of data and information; they would require a newly developed and newly tested system.

In contrast, AI systems that help local councils calculate household benefits may have been built on the (apparently unshakeable) assumption that marriage was between a man and a woman. That knowledge has now changed, however, to permit marriage between any two human adults. Is it reasonable to require a vendor to warn a purchaser that those laws could change too? Such changes seem highly unlikely at present; but in the USA, there have already been attempts by a woman to marry her dog and by a man to marry his laptop, and there has been long-running lobbying from certain religious groups to legalise polygamy.

The US case of Kociemba v Searle [18] found a pharmaceutical manufacturer liable for failing to warn purchasers that use of a particular drug was associated with a pelvic inflammatory disease, even though the product had been passed as "safe and effective" by the Food and Drug Administration. It seems, therefore, that the boundary of where a warning might reasonably be required is indeed dependent on knowledge rather than on regulatory approval.

Mykytyn et al [19] discuss issues of legal liability for AI systems that are linked to identification and selection of human experts. They quote two cases ([20] and [21]) where hospitals were found liable for failing to select physicians with sufficient competence to provide the medical care that they were asked to provide; by analogy, AI developers could also be held liable unless they select experts with sufficient competence in the chosen domain, or warn users that the expert's competence does not extend to other domains where the system might conceivably be used.

The solution proposed by Mykytyn et al. is to use licensed and certified experts. They point out that the standards required by licensing bodies are sometimes used to determine if a professional's performance is up to the level expected [22]. They even suggest that it may be desirable to get the AI system itself licensed. The US Securities and Exchange Commission has been particularly keen on this; it required a stock market recommender system to be registered as a financial adviser [23] and also classified developers of investment advice programs as investment advisors [24].

### 3.2 Limitations of AI systems that do not affect human experts

The key limitation is that AI systems lack general knowledge. Humans carry a great deal of knowledge that is not immediately relevant to a specific task, but that could become relevant. For example, when driving a car, it is advisable to drive slowly when passing a school, especially if there is a line of parked cars outside it, or you know that the school day finishes at about the time when you are driving past. The reason is to avoid the risk of children dashing out from behind parked cars, because a human driver's general knowledge includes the fact that some children have poor road safety skills. An automated car would not know to do this unless it was programmed with a specific rule, or a set of general rules about unusually hazardous locations.

Admittedly, there are occasions when humans fail to apply their general knowledge to recognise a hazardous situation: as one commentator once said, "What is the difference between a belt-driven vacuum cleaner and a Van de Graaff generator? Very little. Never clean your laptop with that type of vacuum cleaner." However, without general knowledge, AI systems have no chance of recognising such situations.

A related issue is that AI systems are notoriously poor at degrading gracefully. This can be seen when considering edge cases (cases where one variable in the case takes an extreme value) or corner cases (multiple variables take extreme values). When human beings are faced with a situation that they previously believed to be very unlikely or impossible, they can usually choose a course of action that has some positive effect on the situation. When AI systems face a situation that they are not programmed for, they generally cannot perform at all.

For example, in the car driving example given at the start of this paper, the (hypothetical) situation where the car refuses to start while a lorry is bearing down on it is an edge case. Furthermore, the car's safety system does not seem to have been designed with city drivers in mind; the car warns drivers to see that their route is safe before making a manoeuvre, but it does not take account of the fact that in a city, a route may only be safe for a short period of time, thus making this type of 'edge' case more common than expected.

As for a corner case, September 26 1983 was the day when a Soviet early-warning satellite indicated first one, then two, then eventually that five US nuclear missiles had been launched. The USSR's standard policy at the time was to retaliate with its own missiles, and it was a time of high political tension between the USA and USSR. The officer in charge had a matter of minutes to decide what to do, and no further information; he chose to consider the message as a false alarm, reasoning that "when people start a war, they don't start it with only five missiles."

It was later discovered that the satellite had mistaken the reflection of sun from clouds as the heat signature of missile launches. The orbit of the satellite was designed to avoid such errors, but on that day (near the Equinox) the location of the satellite, the position of the sun and the location of US missile fields all combined to give five false readings.

If an AI system had been in charge of the Soviet missile launch controls that day, it may well have failed to identify any problem with the satellite, and launched the missiles. It would then have been legally liable for the destruction that followed, although it is unclear whether there would have been any lawyers left to prosecute the case.

A third issue is that AI systems may lack the information that humans use because of poorer quality inputs. This is certainly the case with the car safety system; its only input devices are relatively short range radar detectors, which cannot distinguish between a hedge and a lorry, nor can detect an object that is some distance away but is rapidly approaching. It may be that, should a case come to court regarding an accident 'caused' by these safety systems, the focus will be on how well the AI was programmed to deal with these imprecise inputs.

There is also the issue of non-symbolic information. In the world of knowledge management, it is common to read assertions that human knowledge can never be fully encapsulated in computer systems because it is too intuitive [25]. Kingston [26] argues that this view is largely incorrect because it is based on a poor understanding of the various types of tacit knowledge; but he does allow that non-symbolic information (information based on numbers; shapes; perceptions such as textures; or physiological information e.g. the muscle movements of a ballet dancer), and the skills or knowledge generated from such information, are beyond the scope of nearly all AI systems.

In some domains, this non-symbolic information is crucial: physicians interviewing patients, for example, draw a lot of information from a patient's body language as well as from the patient's words. Some of the criticisms aimed at the UK's current telephone-based diagnostic service, NHS Direct, can be traced back to the medical professional lacking this type of information. In the car-driving example, non-symbolic information might include headlights being flashed by other drivers to communicate messages from one car to another; such information is not crucial but it is important to being a driver who respects others.

## 4 Conclusion

It has been established that the legal liability of AI systems depends on at least three factors:
1. Whether AI is a product or a service. This is ill-defined in law; different commentators offer different views.
2. If a criminal offence is being considered, what *mens rea* is required. It seems unlikely that AI programs will contravene laws that require knowledge that a criminal act was being committed; but it is very possible they might contravene laws for which 'a reasonable man would have known' that a course of action could lead to an offence, and it is almost certain that they could contravene strict liability offences.
3. Whether the limitations of AI systems are communicated to a purchaser. Since AI systems have both general and specific limitations, legal cases on such issues may well be based on the specific wording of any warnings about such limitations.

There is also the question of who should be held liable. It will depend on which of Hallevy's three models apply (perpetrator-by-another; natural-probable-consequence; or direct liability):

- In a perpetrator-by-another offence, the person who instructs the AI system – either the user or the programmer – is likely to be found liable.
- In a natural-or-probable-consequence offence, liability could fall on anyone who might have foreseen the product being used in the way it was; the programmer, the vendor (of a product), or the service provider. The user is less likely to be blamed unless the instructions that came with the product/service spell out the limitations of the system and the possible consequences of misuse in unusual detail.
- AI programs may also be held liable for strict liability offences, in which case the programmer is likely to be found at fault.

However, in all cases where the programmer is deemed liable, there may be further debates whether the fault lies with the programmer; the program designer; the expert who provided the knowledge; or the manager who appointed the inadequate expert, program designer or programmer.